\documentclass[11pt]{article}

\usepackage{amsmath,amssymb,amsfonts,amsthm}
\usepackage{graphicx}
\usepackage{booktabs}
\usepackage{algorithm}
\usepackage{algpseudocode}
\usepackage{bm}
\usepackage{hyperref}
\usepackage{xcolor}
\usepackage{multirow}
\usepackage{subcaption}

\usepackage{orcidlink}
\usepackage{fullpage}

\newtheorem{theorem}{Theorem}

\newcommand{\R}{\mathbb{R}}

\newcommand{\E}{\mathbb{E}}
\newcommand{\inn}[2]{\langle #1,\, #2 \rangle}

\newcommand{\grad}{\nabla}
\newcommand{\DeltaL}{\Delta_{\mathcal{L}}}
\newcommand{\thetastar}{\theta^*}
\newcommand{\fprime}{f^{\prime}}
\newcommand{\etaprime}{\eta^{\prime}}
\newcommand{\Hsq}{H^{1/2}}
\newcommand{\Hisq}{H^{-1/2}}

\algnewcommand\RETURN{\State\algorithmicreturn}

\title{Neural Legendre--Fenchel transform\\ with Hessian Preconditioning}

\author{
Basile Plus-Gourdon\\
\'Ecole Normale Sup\'erieure\\
Paris-Saclay, France
\and 
Frank Nielsen~\orcidlink{0000-0001-5728-0726}\\
Sony Computer Science Laboratories Inc.\\
Tokyo, Japan
}

\date{}

\begin{document}

\maketitle

\begin{abstract}
The Legendre–-Fenchel (LF) transform is a fundamental tool in convex analysis and machine learning that maps lower semi-continuous functions to their convex conjugates. 
In practice, when closed-form formula are not available for expressing convex conjugates of given functions, one must approximate them  using various techniques.  
One recent such versatile numerical method is the deep Legendre transform method which relies on neural networks although it  remains challenging particularly for tackling ill-conditioned functions.
This work builds on the reformulation of the LF transform  as a projective polarity, leading to a neural training objective expressed as a matrix-valued polar divergence.
 A notable property of this framework is its affine invariance: Namely, any affine deformation of the input function induces an equivalent conjugation problem under a corresponding reparameterization.
In this letter, we leverage this affine invariance to introduce a Hessian-based preconditioning strategy which improves conditioning and generally leads to
faster convergence and more accurate conjugations. 
Specifically, we apply an affine deformation around a minimizer so that the second-order Taylor approximation of the function coincides with the canonical paraboloid, whose conjugation map is the identity. A residual network initialized near the identity can then learn this simplified mapping, while the original conjugation map is recovered through the inverse deformation.
The proposed preconditioning incurs only a modest computational overhead, consisting of a single eigendecomposition during initialization and two matrix–vector multiplications per query. Experiments on a diverse set of convex functions, including high-dimensional benchmarks, demonstrate improved convergence rates and enhanced numerical accuracy of the conjugation, with particularly significant gains for ill-conditioned problems.
Finally, we discuss the scope of applicability of our proposed method and highlight several of its limitations.
\end{abstract}


\section{Introduction}

The Legendre--Fenchel (LF) transform~\cite{Bauschke2012} $f^*(\eta)=\sup_{\theta\in\Theta} \{ \inn{\theta}{\eta}-f(\theta) \}$ of a $n$-variate function $f:\Theta\subset\mathbb{R}^n\rightarrow\mathbb{R}$
is an essential tool in convex analysis. 
When $f$ is of Legendre-type~\cite{Rockafellar1967}, the convex conjugate is expressed as $f^*(\eta)=\inn{\theta^*}{\eta}-f(\theta^*)$ for $\theta^*=(\nabla f)^{-1}(\eta)$. This formula was first reported by Adrien-Marie Legendre (1752-1833) when $n=1$ and was later extended in a more general setting by Werner Fenchel (1905-1988) in arbitrary dimension.
However, when $f$ is not available in closed-form or is computationally intractable (e.g., log-partition functions of discrete exponential families~\cite{rinaldo2009geometry}) or when $\nabla f$ is not invertible in closed form, one has to numerically evaluate or estimate the conjugation function.

The LF transform is fundamental in many scientific areas like in 
Hamiltonian mechanics, in information geometry, and in machine learning among others.
For instance, solving the Kantorovich formulation of the
Wasserstein-$2$ optimal transport (OT) problem requires evaluating the
conjugate $f^*$~\cite{Villani2009} as a subtask, motivating scalable
neural approximations.

Several works have investigated neural approximations of the Legendre--Fenchel transform. Early approaches in optimal transport parameterized the potential with input-convex neural networks (ICNNs) and recovered the conjugate through an inner optimization procedure \cite{Taghvaei2019,Makkuva2020}. Subsequent works introduced amortized neural approximations of the conjugation map, either by directly predicting the optimizer of the Fenchel problem or by learning the conjugate function itself \cite{korotin2020wasserstein2generativenetworks,korotin2021neuraloptimaltransportsolvers,amos2025amortizingconvexconjugatesoptimal}.
 Recently, Deep Legendre Transform~\cite{minabutdinov2026deeplegendretransform} (DLT) proposed learning the mapping $\eta\mapsto f^*(\eta)$ directly from samples generated via $\eta=\nabla f(\theta)$, enabling efficient approximation of conjugate values in higher dimensions. In contrast, our method learns the conjugation map $\eta\mapsto\theta^*(\eta)=\nabla f^*(\eta)$ directly and does not require evaluating $\nabla f$ at inference time. 

In this work we learn the conjugation map
$\eta\mapsto\thetastar(\eta)$ by minimizing the Fenchel-Young divergence, which was
recently given a matrix interpretation through projective polarity
\cite{Nielsen2026}.
In that framework, the LF transform of a $n$-variate function is the boundary of a polar set induced by
a $(n{+}2)\times(n{+}2)$ matrix~$C_\mathcal{L}= \begin{bmatrix}
            -I_n & 0 & 0 \\
            0 & 0 & 1 \\
            0 & 1 & 0
        \end{bmatrix}$, and the training loss equals
$D_A(a{:}b)=[a]^\top C_\mathcal{L}[b]$ where $[a]$ and $[b]$ are
homogeneous-coordinate representations of points on $\operatorname{graph}(f)$
and $\operatorname{graph}(f^*)$, respectively.
Crucially, Theorems~1 and~2 of~\cite{Nielsen2026} establish that any affine
deformation $f\mapsto f\circ T^{-1}$ is equivalent to a change of the polarity
matrix $C$, leaving the structure of the learning problem unchanged.

\textit{Contribution.} We leverage this affine invariance in a concrete
algorithm.
We deform $f$ by the inverse square root of its Hessian at the minimizer,
producing a preconditioned function whose local geometry matches the canonical
paraboloid.
Because half of the paraboloid is self-conjugate ($Q^*=Q$ with $Q(\theta)=\frac{\theta^2}{2}$) and the conjugation map is the
identity, the residual network, initialized near identity, starts in an
excellent basin.
Our contributions are summarized as follows:
\begin{itemize}
  \item A theoretically grounded Hessian preconditioning procedure
        (\S\ref{sec:method}) directly motivated by the polarity framework.
  \item An efficient inference scheme: the inverse deformation costs two
        matrix-vector products per query.
  \item Empirical evidence across several convex function classes and
        dimensions $n\in\{4,10,20,50\}$ that preconditioning consistently
        improves convergence and conjugation quality.
\end{itemize}

\section{Preliminaries}
\label{sec:background}

\subsection{Neural Legendre--Fenchel Conjugation}
\label{ssec:neural_lf}

Let $f:\Theta\subset\R^n\to\R$ be a closed convex function of
Legendre type~\cite{Rockafellar1967}, so that its conjugate
$f^*:\mathcal{H}\subset\R^n\to\R$ satisfies $\grad f^*=(\grad f)^{-1}$.
The \emph{conjugation map} $\thetastar:\mathcal{H}\to\Theta$ (i.e., inverse gradient map) is defined by
\begin{equation}
  \thetastar(\eta)
  = \operatorname*{argmax}_{\theta\in\Theta}
    \{\inn{\theta}{\eta}-f(\theta)\}
  = \grad f^*(\eta).
\end{equation}
We parameterize a \emph{residual network}
$\theta_\varphi(\eta)=\eta+\operatorname{MLP}_\varphi(\eta)$,
with the multilayer perceptron (MLP) output layer zero-initialized so that
$\theta_\varphi\equiv\operatorname{id}$ at initialization.
Training minimizes the \emph{Fenchel-Young (LF) loss}~\cite{Blondel2020}:
\begin{equation}
  \mathcal{L}(\varphi)
  = \E_\eta\bigl[f(\theta_\varphi(\eta))-\inn{\theta_\varphi(\eta)}{\eta}\bigr].
  \label{eq:lf_loss}
\end{equation}

Once trained, $f^*(\eta)$ can be recovered approximately as
$\hat f^*(\eta)=\inn{\theta_\varphi(\eta)}{\eta}-f(\theta_\varphi(\eta))$.

\subsection{Legendre Polarity and Affine Invariance}
\label{ssec:polarity}

In~\cite{fenchel2013conjugate,Nielsen2026}  the LF transform is interpreted as a
\emph{projective polarity}.
Points $(\theta,f(\theta))$ on $\operatorname{graph}(f)$ are embedded in the projective
space $\mathbb{P}^{n+1}$ via homogeneous coordinates
$[a]=[\theta^\top\;f(\theta)\;1]^\top\in\R^{n+2}$.
The \emph{Legendre polarity} $\DeltaL$ is induced by the symmetric matrix
\begin{equation}
  C_\mathcal{L} =
  \begin{bmatrix}
    -I_n & 0 & 0 \\
    0    & 0 & 1 \\
    0    & 1 & 0
  \end{bmatrix}.
\end{equation}
The polar divergence between a primal point
$[a]=[\theta^\top\;f(\theta)\;1]^\top$ and a dual point
$[b]=[\eta^\top\;f^*(\eta)\;1]^\top$ is
\begin{eqnarray}
  D_A(a:b) &=& [a]^\top C_\mathcal{L}[b],\nonumber\\
	&=& f(\theta)+f^*(\eta)-\inn{\theta}{\eta}=:Y_f(\theta:\eta),
  \label{eq:polar_div}
\end{eqnarray}
which precisely recovers the Fenchel-Young divergence.
Thus minimizing~\eqref{eq:lf_loss} is equivalent to minimizing the polar
divergence~\eqref{eq:polar_div} over the learned map.

The central invariance result of~\cite{Nielsen2026} is:

\begin{theorem}[Affine invariance, Theorems~1--2 of~\protect\cite{Nielsen2026}]
\label{thm:affine_inv}
Let $S:\R^{n+1}\to\R^{n+1}$ be an invertible affine deformation with
matrix $M_S\in\mathrm{GL}(n{+}2)$.  Then we have
\begin{equation}
  \DeltaL(S(A)) = \DeltaL(S(A))
  ,
\end{equation}
\begin{equation}
    \partial\DeltaL(S(\operatorname{graph}(f)))
  = \operatorname{graph}\left((f\circ T^{-1})^*\right)
\end{equation}
where $T$ is the dual deformation satisfying $M_T = C_\mathcal{L}M_S^{-\top}C_\mathcal{L}$.
\end{theorem}

In plain words: The Legendre conjugate of an affinely deformed function is the
affinely deformed conjugate of the original function.
This implies that \emph{learning to conjugate $f$ is equivalent to learning
to conjugate the deformed $f\circ S^{-1}$}, with a corresponding linear
transformation of the output.

\section{Hessian Preconditioning}
\label{sec:method}

\subsection{Motivation}

The quadratic paraboloid $Q(\theta)=\frac{1}{2}\|\theta\|^2$ is its own
Legendre conjugate: $Q^*=Q$~\cite{Bauschke2012}.
Its conjugation map is the identity, $\thetastar(\eta)=\eta$, which the
zero-initialized residual network can fit without any training.
This suggests that if $f$ locally resembles $Q$ near its minimizer, learning
will be substantially easier.

Consider the second-order Taylor expansion at a minimizer $\theta_0$,
\begin{equation}
  f(\theta)
  \approx f(\theta_0)+\tfrac{1}{2}(\theta-\theta_0)^\top H(\theta-\theta_0),
  \quad H=\nabla^2 f(\theta_0).
  \label{eq:taylor}
\end{equation}
Under the change of variable $z=\Hsq(\theta-\theta_0)$ this becomes
$f \approx f(\theta_0)+\frac{1}{2}\|z\|^2$.
Theorem~\ref{thm:affine_inv} guarantees that the conjugation problem under
this deformation has the same structure, so the learning task in $z$-space is
approximately that of the canonical paraboloid.

\subsection{Preconditioned Function and Dual Parameterization}

Let $H=V\Lambda V^\top$ be the eigendecomposition of the Hessian, with
$\Hisq=V\Lambda^{-1/2}V^\top$.
We define another function $f'$ as follows:
\begin{align}
  \fprime(z) &= f\!\bigl(\Hisq z+\theta_0\bigr)-f(\theta_0),
  \label{eq:fprime}\\
  \etaprime  &= \eta\Hisq\quad\text{(row-vector convention)}.
  \label{eq:etaprime}
\end{align}
By a direct change of variables in the supremum:
\begin{align}
  (f')^*(\etaprime)
  &= f^*(\Hsq\etaprime) - \inn{\theta_0}{\Hsq\etaprime} + f(\theta_0),
  \label{eq:fprime_star}\\
  (\fprime)^{*\prime}(\etaprime)
  &= \Hisq\, \thetastar(\Hsq\etaprime) - \theta_0
    \;=:\; z^*(\etaprime).
  \label{eq:zstar}
\end{align}
Hence the original conjugation map is recovered as
\begin{equation}
  \thetastar(\eta) = \Hisq z^*(\eta\Hisq)+\theta_0.
  \label{eq:decode}
\end{equation}
Near the minimizer, $\fprime(z)\approx\frac{1}{2}\|z\|^2$, so
$z^*(\etaprime)\approx\etaprime$, i.e., the learning problem is almost trivial in a
neighborhood of the origin.

\subsection{Algorithms}

Algorithms~\ref{alg:setup}--\ref{alg:infer} describe the full procedure.
The setup cost is dominated by (i) minimizing $f$ to find $\theta_0$ and (ii)
the eigendecomposition of $H\in\R^{n\times n}$, both of which are
operations computed only one time.
At inference, the overhead over the baseline is exactly two matrix-vector
products (lines~\ref{line:mv1} and~\ref{line:mv2} of
Algorithm~\ref{alg:infer}).

\begin{algorithm}[tb]
\caption{Hessian Preconditioning Setup}
\label{alg:setup}
\begin{algorithmic}[1]
  \Require convex function $f:\R^n\to\R$, primal range $[\theta_\ell,\theta_u]$
  \State $\theta_0\leftarrow\operatorname*{argmin}_\theta f(\theta)$
         
  \State $H\leftarrow\nabla^2 f(\theta_0)$
         
  \State $\Hisq\leftarrow V\Lambda^{-1/2}V^\top$
        
  \State $\fprime(z)\leftarrow f(\Hisq z+\theta_0)-f(\theta_0)$
  \State compute $\eta'_{\min},\eta'_{\max}$ from $\grad f$ over $[\theta_\ell,\theta_u]$;
         transform by $\Hisq$
  \RETURN\ $\Hisq,\;\theta_0,\;\fprime,\;\eta'_{\min},\;\eta'_{\max}$
\end{algorithmic}
\end{algorithm}

\begin{algorithm}[tb]
\caption{Preconditioned Training}
\label{alg:train}
\begin{algorithmic}[1]
  \Require $\fprime$, $\eta'_{\min},\eta'_{\max}$, steps $T$
  \State Initialize $\theta_\varphi(z)=z+\operatorname{MLP}_\varphi(z)$,
         \;$\varphi$~zero-initialized
  \For{$t=1,\ldots,T$}
    \State sample $\eta'\sim\mathcal{U}[\eta'_{\min},\eta'_{\max}]$
    \State $z\leftarrow\theta_\varphi(\eta')$
    \State $\mathcal{L}\leftarrow\E\!\left[\fprime(z)-\inn{z}{\eta'}\right]$
    \State update $\varphi$ with one Adam step on $\nabla_\varphi\mathcal{L}$
  \EndFor
  \RETURN\ $\theta_\varphi$
\end{algorithmic}
\end{algorithm}

\begin{algorithm}[tb]
\caption{Preconditioned Inference}
\label{alg:infer}
\begin{algorithmic}[1]
  \Require query $\eta$, network $\theta_\varphi$, $\Hisq$, $\theta_0$, function $f$
  \State $\etaprime\leftarrow\eta\,\Hisq$\label{line:mv1}
         \quad\Comment{dual deformation}
  \State $z\leftarrow\theta_\varphi(\etaprime)$
  \State $\thetastar(\eta)\leftarrow z\,\Hisq+\theta_0$\label{line:mv2}
         \quad\Comment{inverse primal deformation; cf.~\eqref{eq:decode}}
  \State $\hat f^*(\eta)\leftarrow\inn{\thetastar(\eta)}{\eta}-f(\thetastar(\eta))$
  \RETURN\ $\thetastar(\eta),\;\hat f^*(\eta)$
\end{algorithmic}
\end{algorithm}

\section{Experiments}
\label{sec:experiments}

\subsection{Experimental Setup}

\textbf{Architecture.}
Both baseline and preconditioned models use a residual network
$\theta_\varphi(\eta)=\eta+\operatorname{MLP}(\eta)$: 3 hidden layers of
width~64, ReLU activations, output layer zero-initialized.
We train with Adam ($\mathrm{lr}=10^{-3}$, batch size 512) and minimize
loss~\eqref{eq:lf_loss}.

\textbf{Ground truth.}
We compute $\thetastar(\eta)$ by Adam optimization of
$-\inn{\theta}{\eta}+f(\theta)$ (1500 steps, $\mathrm{lr}=0.05$,
gradient clipping), which gives high-quality numerical conjugates.

\textbf{Metrics.}
We report $\theta$ relative RMSE $\frac{\|\theta_\varphi(\eta)-\thetastar(\eta)\|_2}{\|\theta^*(\eta)\|_2}$
(conjugation quality) and $f^*$ relative RMSE $\frac{\|\hat f^*(\eta)-f^*_\mathrm{GT}(\eta)\|_2}{\|f^*_{\mathrm{GT}}(\eta)\|_2}$
(conjugate value quality), where $f^*(\eta)$ is re-constructed from the predicted conjugate value $\theta_\varphi(\eta)$.

\subsection{Preconditioning Effectiveness}

\begin{figure}[tb]
  \centering
  \includegraphics[width=0.95\linewidth]{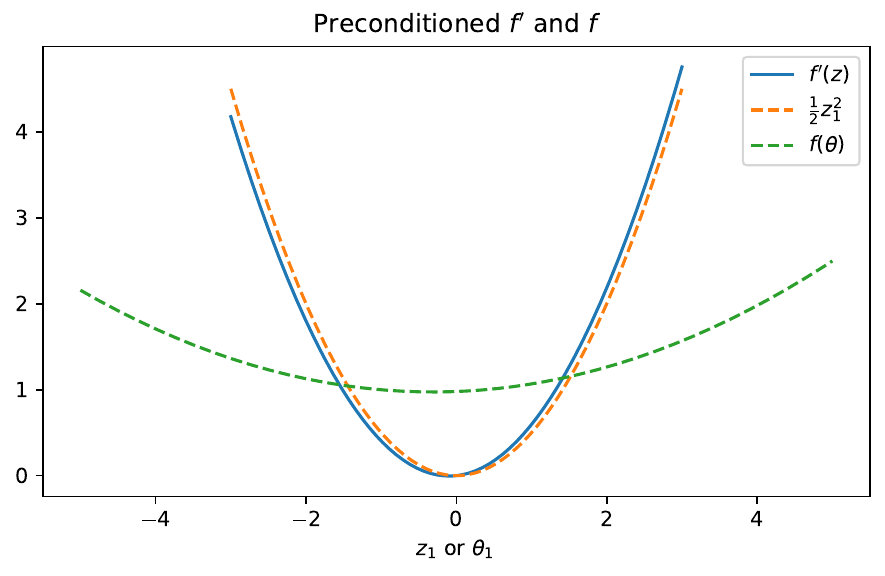}
  \caption{Effectiveness of the preconditioning  on a randomly initialised ICNN.
           After the Hessian transformation, $\fprime(z)$
           closely matches the canonical paraboloid $\frac{1}{2}z_1^2$
           (blue), while the original $f(\theta)$ has a
           very different scale.}
  \label{fig:precond_check}
\end{figure}

Figure~\ref{fig:precond_check} illustrates the preconditioning for a
\emph{randomly initialised ICNN}.
After the Hessian transformation, $\fprime$ closely follows
$\frac{1}{2}z_1^2$ near the origin, while the original $f$ has a very
different curvature profile.
This near-paraboloidal shape means the conjugation map in $z$-coordinates
is close to the identity, and the residual network converges from its
zero initialization with minimal gradient effort.

\subsection{Training and Conjugation Comparison}

\begin{figure}[tb]
  \centering
  \begin{subfigure}[b]{0.49\linewidth}
    \centering
    \includegraphics[width=\linewidth]{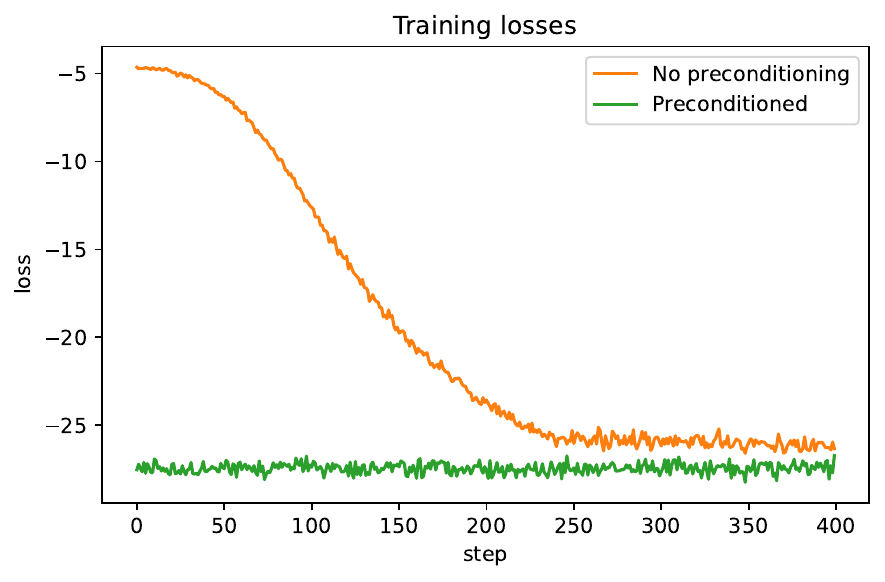}
    \caption{Training losses.}
    \label{fig:losses}
  \end{subfigure}
  \hfill
  \begin{subfigure}[b]{0.49\linewidth}
    \centering
    \includegraphics[width=\linewidth]{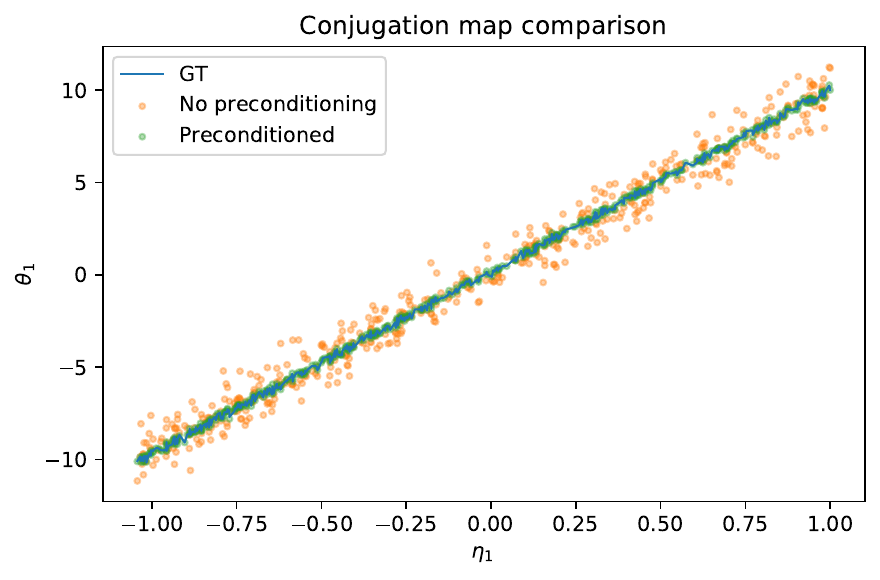}
    \caption{Conjugation map $\theta_1\leftrightarrow\eta_1$.}
    \label{fig:conjugate}
  \end{subfigure}
  \caption{Training loss (left) and conjugation map quality (right)
           for the randomly initialised ICNN.
           Preconditioned (green) achieves lower loss from the outset and
           tracks the GT (blue) more closely than the baseline (orange).}
  \label{fig:losses_conjugate}
\end{figure}

\begin{figure}[tb]
  \centering
  \includegraphics[width=0.95\linewidth]{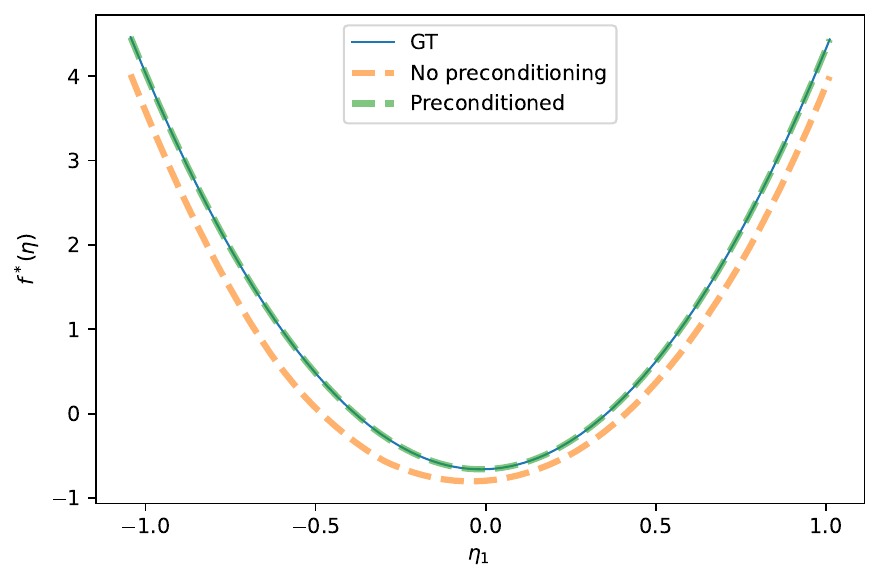}
  \caption{Reconstructed conjugate $\hat f^*(\eta)$ vs.\ ground truth for a
           randomly initialised ICNN.
           Preconditioned (green dashed) aligns with the GT (blue), while
           the baseline (orange dashed) shows noticeable deviation.}
  \label{fig:fstar}
\end{figure}

Figures~\ref{fig:losses_conjugate} and~\ref{fig:fstar} compare training
dynamics and output quality (here $n=4$).
The preconditioned model converges to a stable low loss from the very first
steps, while the baseline undergoes a prolonged transient before settling
at a noticeably higher value (Figure~\ref{fig:losses}).
On the $\theta$-map (Figure~\ref{fig:conjugate}), the preconditioned
prediction tightly tracks the ground truth (GT), whereas the baseline exhibits
visible scatter.
The recovered conjugate function $\hat f^*$ (Figure~\ref{fig:fstar})
further confirms the advantage: the preconditioned model aligns nearly
perfectly with the ground truth, while the baseline shows systematic bias.
Similar improvements are observed across all function families tested
(see Appendix and Table~\ref{tab:benchmark}).

\subsection{Multi-Function Quantitative Benchmark}

Table \ref{tab:benchmark} reports $\theta$-RelRMSE and $f^*$-RelRMSE for four
function families across $n\in\{10,20,50\}$.
The functions are:
\begin{enumerate}
  \item \textbf{Scaled quadratic}: $f(\theta)=\frac{1}{2}\theta^\top Q\theta$,
        $Q$ random symmetric positive definite (SPD) matrix, $\kappa(Q)\approx 100$
  \item \textbf{Quartic mix}: $f(\theta)=\frac{1}{2}\|\theta\|^2
        +\frac{1}{2}\|\theta-\mathbf{1}\|^4$
  \item \textbf{Sum cosh}: $f(\theta)=\sum_i\cosh(0.7\theta_i-1)+2\theta_i$
  \item \textbf{Exp-quad}: $f(\theta)=\sum_i e^{0.5(\theta_i-1)}
        +\frac{1}{2}\theta_i^2-\theta_i$
    \item \textbf{Log-Sum-Exp (regularized)}: $f(\theta)=\log\sum_i e^{\theta_i} + 0.1\|\theta\|^2 $
\end{enumerate}
Preconditioning provides consistent gains across all functions and
dimensions.
The improvement is most dramatic for the ill-conditioned scaled quadratic,
where the preconditioning is specially efficient.

Note that the training fails for some functions, for instance the quartic $f(\theta)=\frac{1}{2}\|\theta\|^2+\frac{1}{2}\|\theta-\mathbf{1}\|^4$. This is due to really high values of $f(\theta)$ that induces a loss divergence toward infinity.

\begin{figure*}[p]
\centering

\textbf{Quartic Mix}\par\vspace{1mm}

\begin{subfigure}[b]{0.24\textwidth}
    \includegraphics[width=\linewidth]{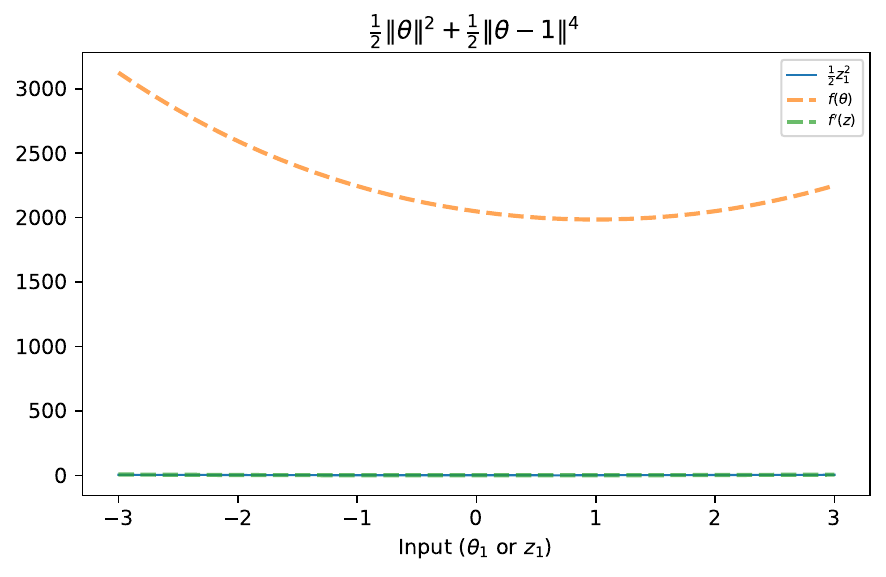}
\end{subfigure}
\hfill
\begin{subfigure}[b]{0.24\textwidth}
    \includegraphics[width=\linewidth]{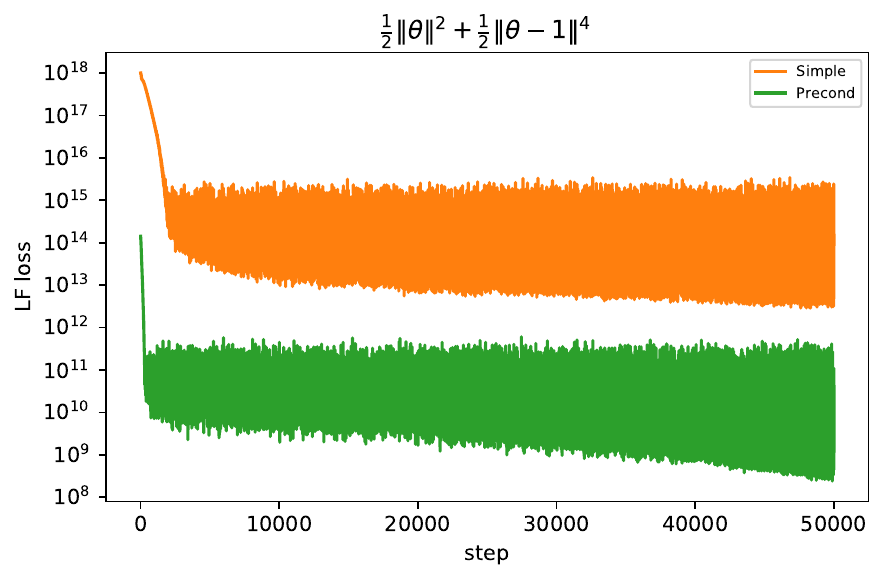}
\end{subfigure}
\hfill
\begin{subfigure}[b]{0.24\textwidth}
    \includegraphics[width=\linewidth]{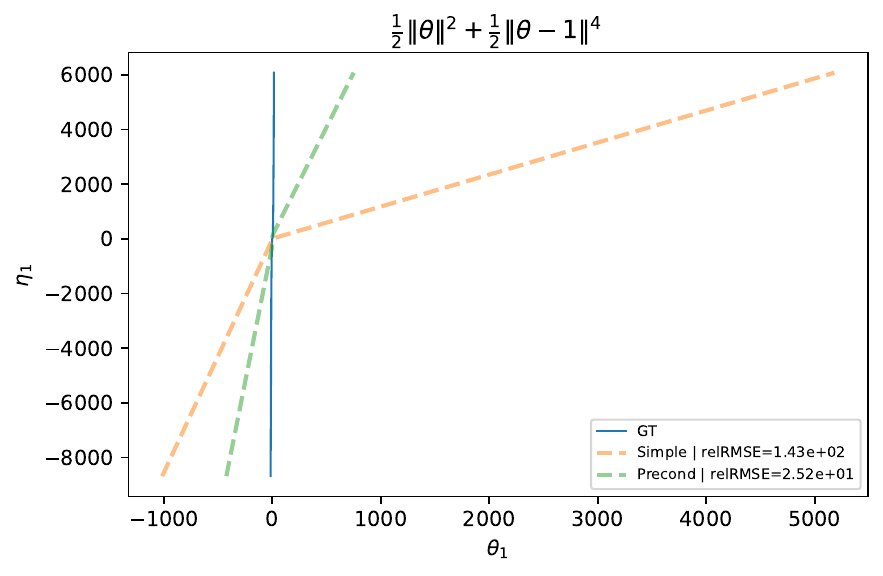}
\end{subfigure}
\hfill
\begin{subfigure}[b]{0.24\textwidth}
    \includegraphics[width=\linewidth]{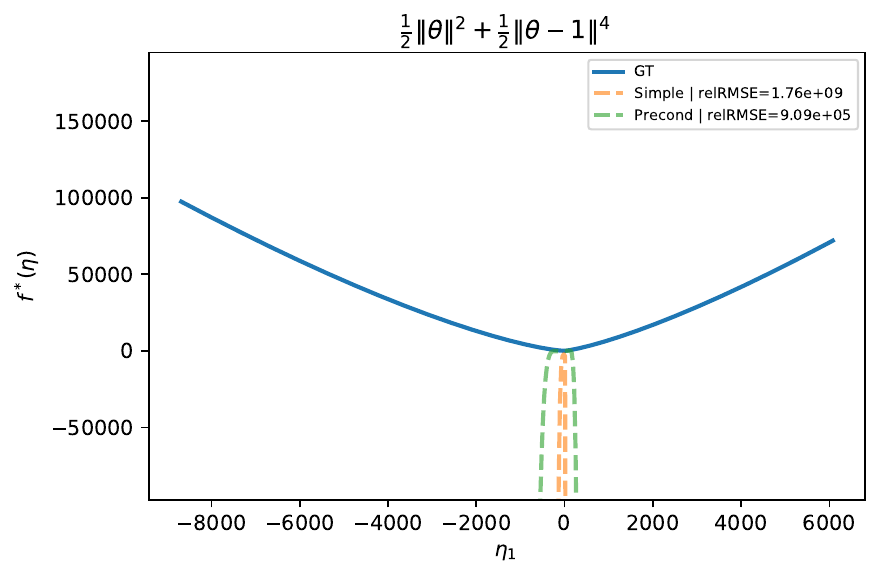}
\end{subfigure}

\vspace{2mm}

\textbf{Sum Cosh}\par\vspace{1mm}

\begin{subfigure}[b]{0.24\textwidth}
    \includegraphics[width=\linewidth]{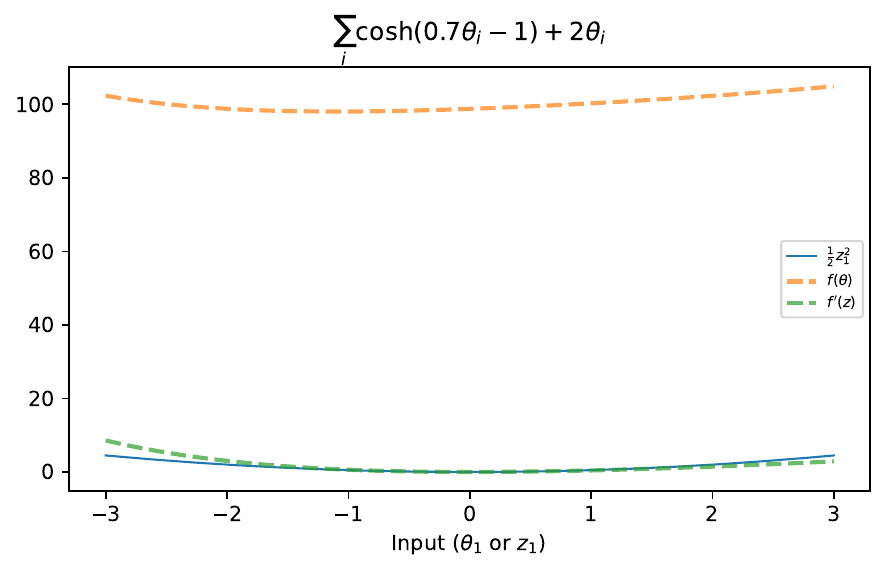}
\end{subfigure}
\hfill
\begin{subfigure}[b]{0.24\textwidth}
    \includegraphics[width=\linewidth]{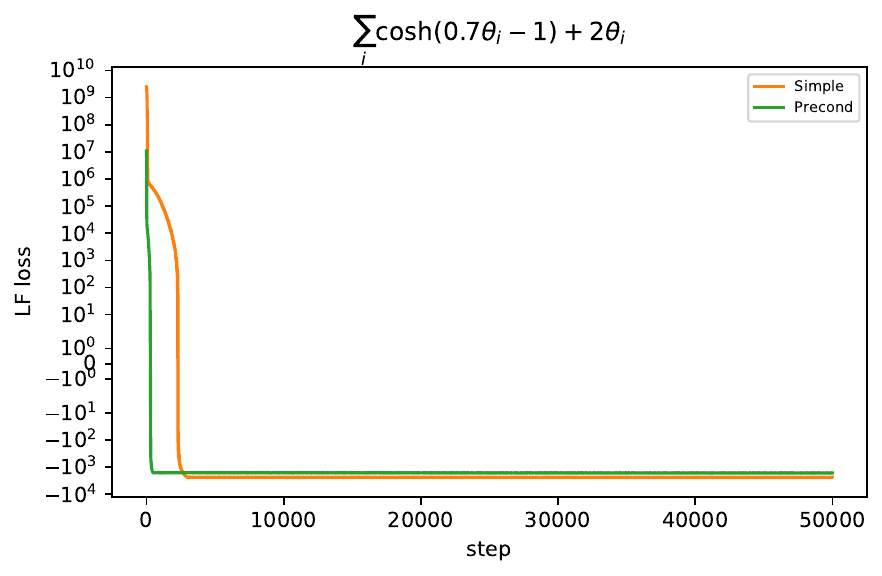}
\end{subfigure}
\hfill
\begin{subfigure}[b]{0.24\textwidth}
    \includegraphics[width=\linewidth]{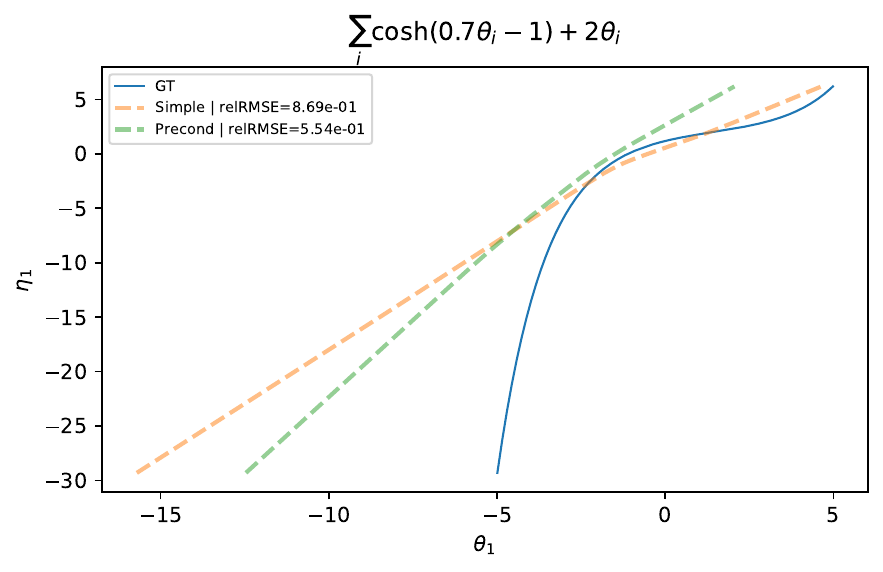}
\end{subfigure}
\hfill
\begin{subfigure}[b]{0.24\textwidth}
    \includegraphics[width=\linewidth]{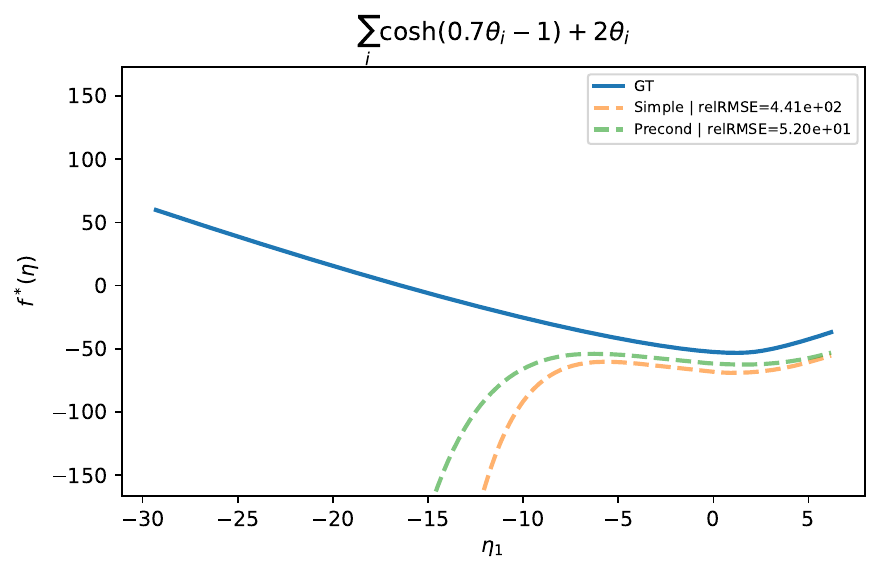}
\end{subfigure}

\vspace{2mm}

\textbf{Exponential + Quadratic}\par\vspace{1mm}

\begin{subfigure}[b]{0.24\textwidth}
    \includegraphics[width=\linewidth]{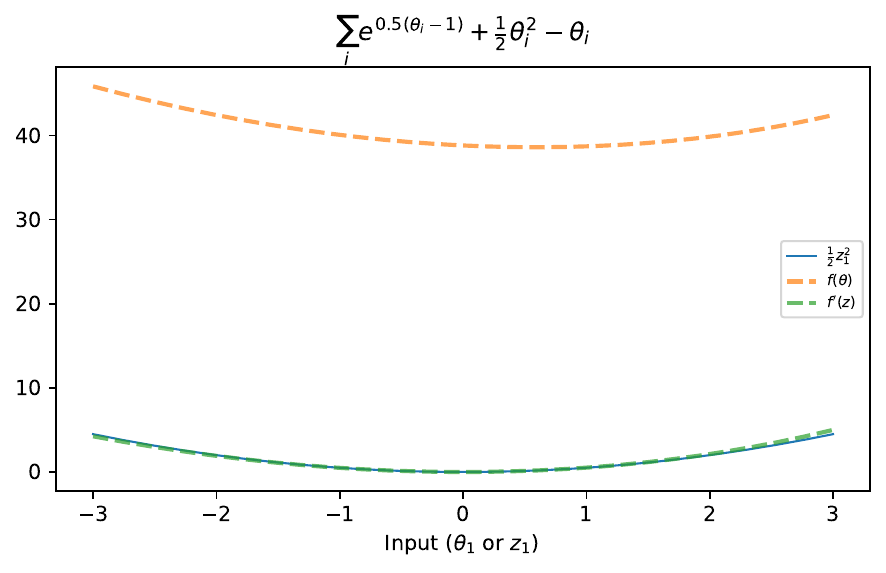}
\end{subfigure}
\hfill
\begin{subfigure}[b]{0.24\textwidth}
    \includegraphics[width=\linewidth]{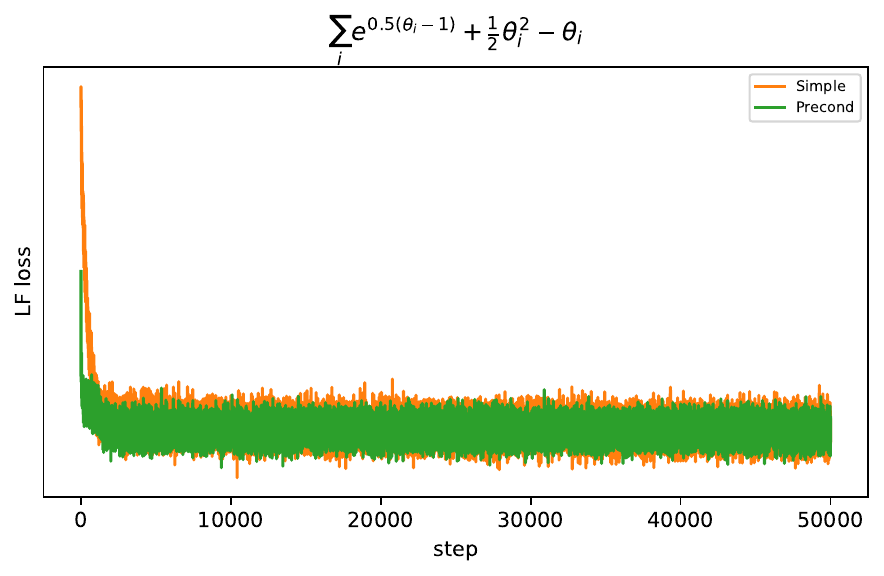}
\end{subfigure}
\hfill
\begin{subfigure}[b]{0.24\textwidth}
    \includegraphics[width=\linewidth]{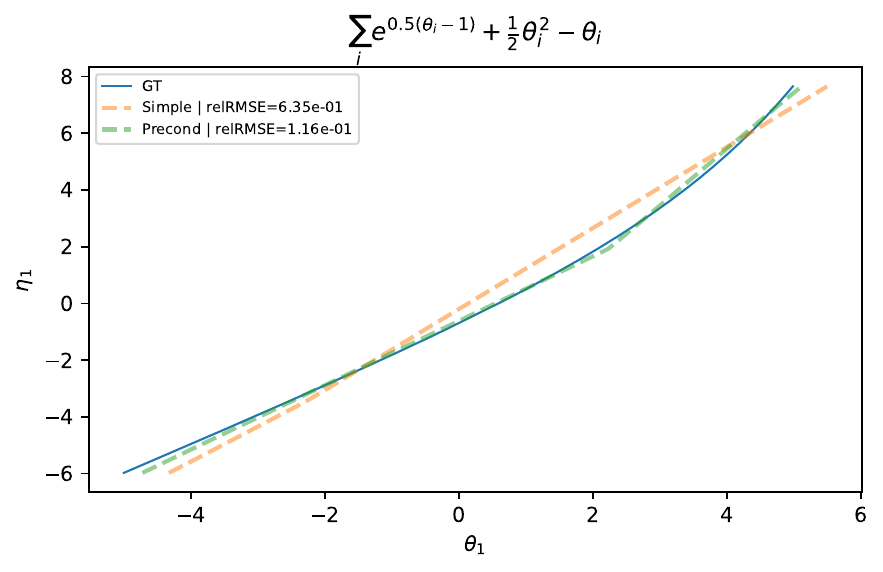}
\end{subfigure}
\hfill
\begin{subfigure}[b]{0.24\textwidth}
    \includegraphics[width=\linewidth]{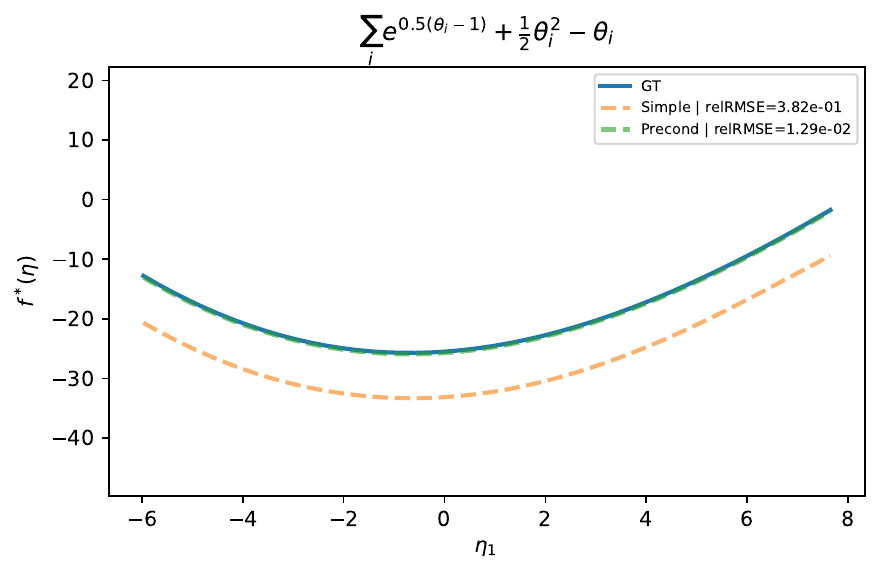}
\end{subfigure}

\vspace{2mm}

\textbf{Log-Sum-Exp (regularized)}\par\vspace{1mm}

\begin{subfigure}[b]{0.24\textwidth}
    \includegraphics[width=\linewidth]{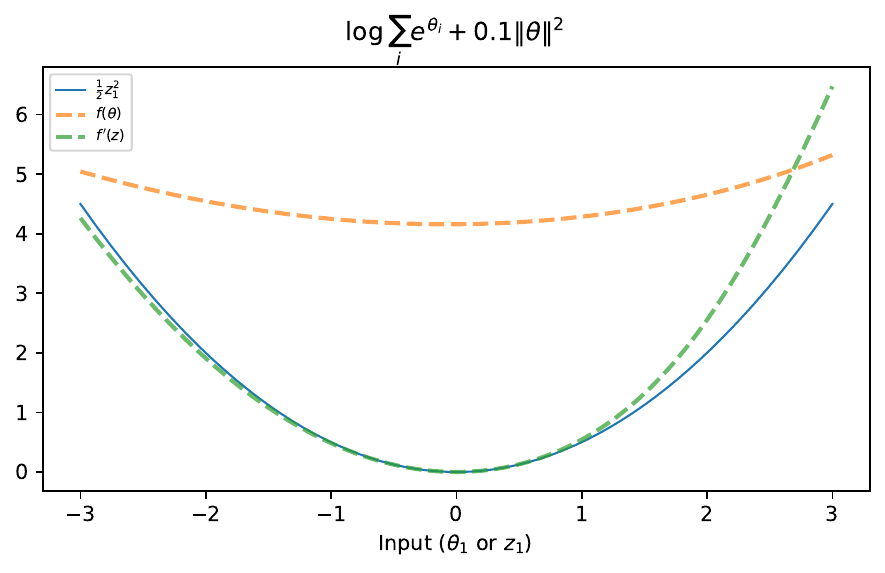}
\end{subfigure}
\hfill
\begin{subfigure}[b]{0.24\textwidth}
    \includegraphics[width=\linewidth]{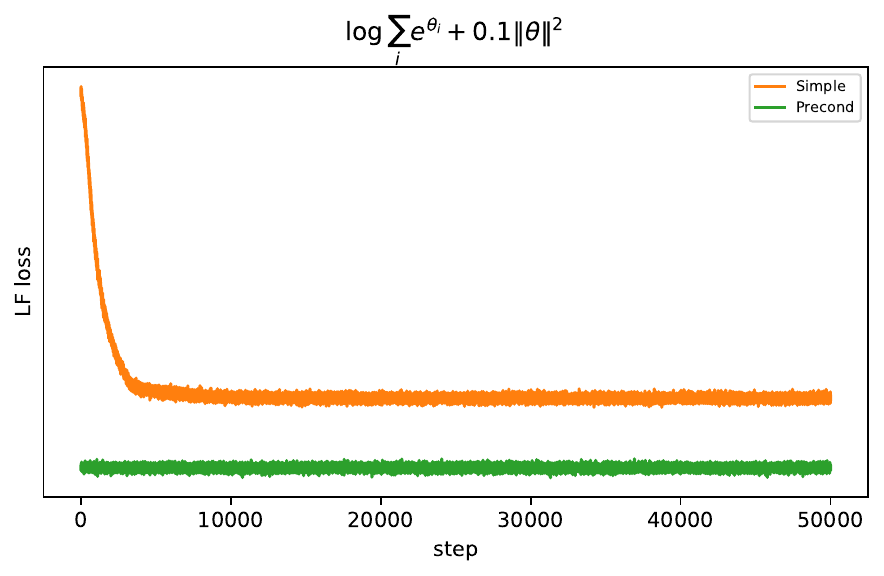}
\end{subfigure}
\hfill
\begin{subfigure}[b]{0.24\textwidth}
    \includegraphics[width=\linewidth]{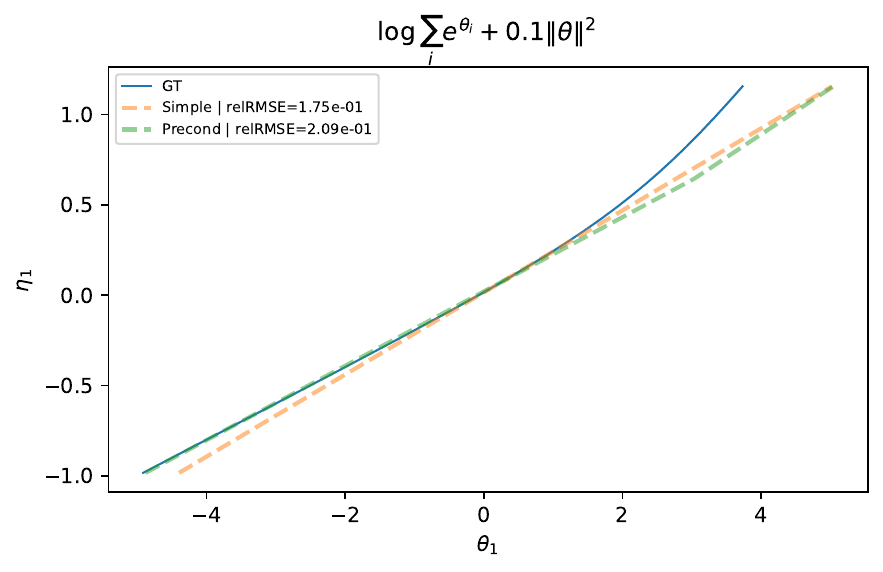}
\end{subfigure}
\hfill
\begin{subfigure}[b]{0.24\textwidth}
    \includegraphics[width=\linewidth]{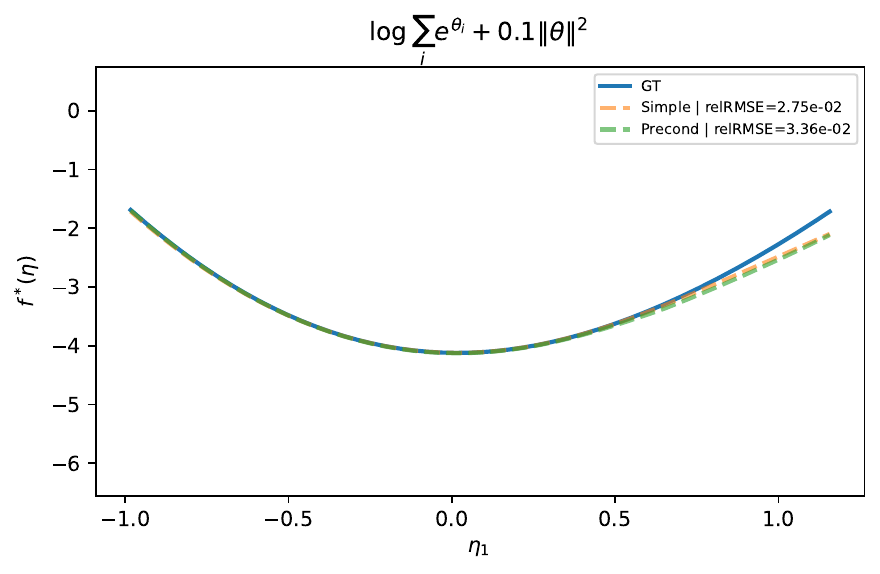}
\end{subfigure}

\vspace{2mm}

\textbf{Scaled Quadratic $\times 50$}\par\vspace{1mm}

\begin{subfigure}[b]{0.24\textwidth}
    \includegraphics[width=\linewidth]{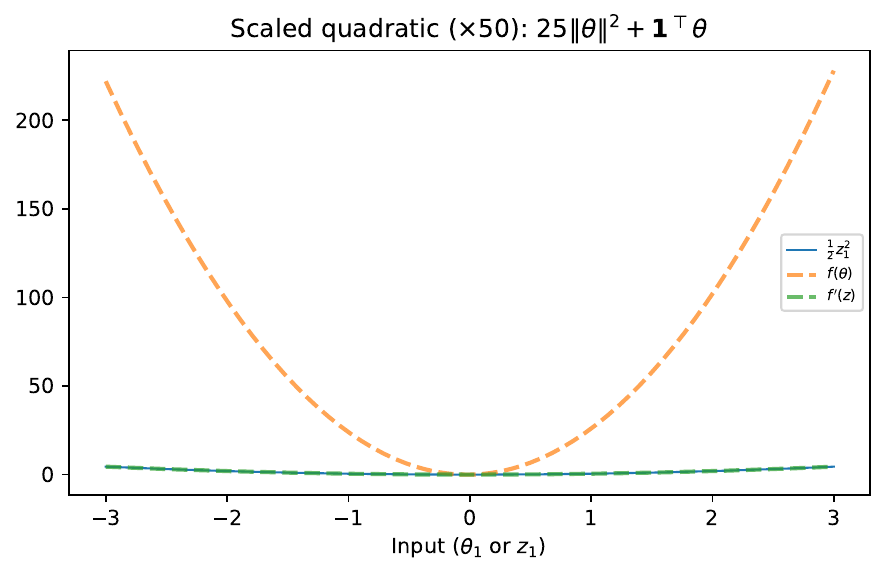}
\end{subfigure}
\hfill
\begin{subfigure}[b]{0.24\textwidth}
    \includegraphics[width=\linewidth]{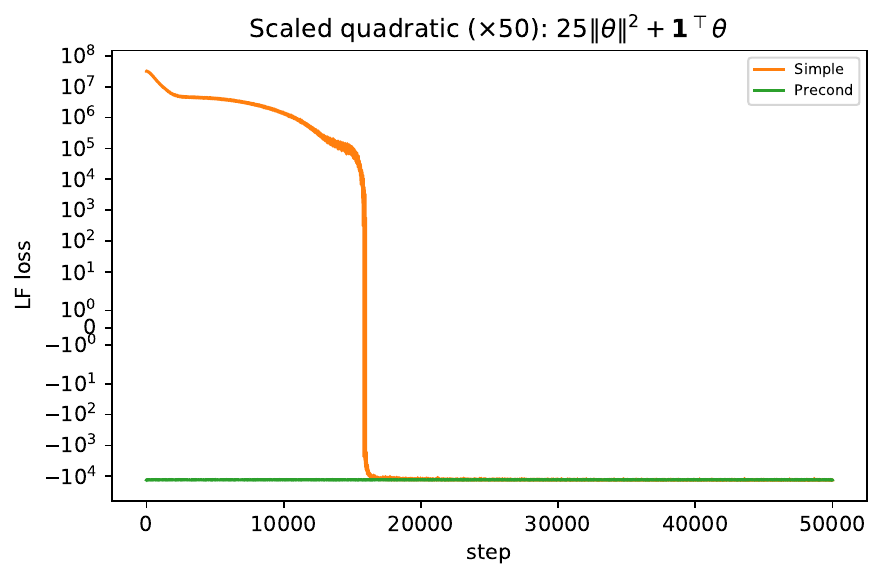}
\end{subfigure}
\hfill
\begin{subfigure}[b]{0.24\textwidth}
    \includegraphics[width=\linewidth]{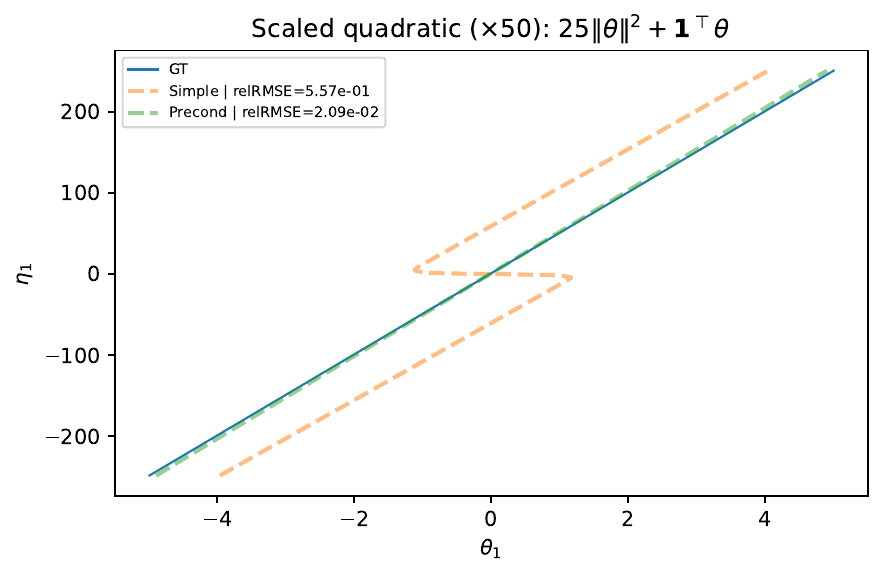}
\end{subfigure}
\hfill
\begin{subfigure}[b]{0.24\textwidth}
    \includegraphics[width=\linewidth]{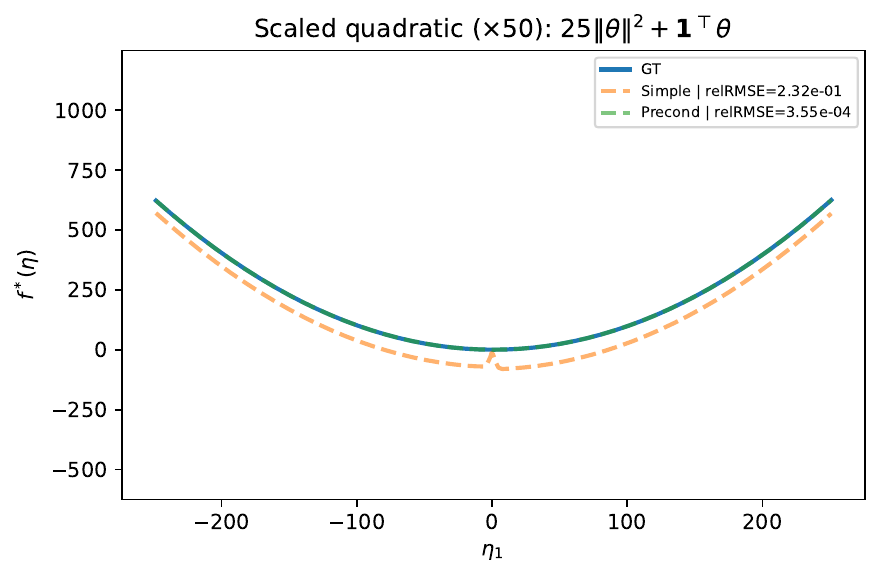}
\end{subfigure}

\caption{
Additional experiments in dimension $n=64$.
For each function family, we show
(from left to right)
the effect of Hessian preconditioning,
the training loss,
the learned conjugation map $\theta^*(\eta)$,
and the reconstructed convex conjugate $f^*(\eta)$.
Preconditioning consistently improves conditioning and generally leads to
faster convergence and more accurate recovery of both the conjugation map
and the conjugate function.
}
\label{fig:appendix_all}
\end{figure*}

It is also important to note (Figure \ref{fig:losses} and \ref{fig:appendix_all}) that in every cases, in addition to more accuracy, the preconditionning also greatly improves the learning speed of the network. See for instance the scaled quadratic figure \ref{fig:appendix_all} where the baseline method requires more than 15000 training steps, whereas the preconditioned conjugation is learned in just a few training steps.

\begin{table*}[tb]
\caption{Benchmark: $\theta$-RMSE and $f^*$-RMSE (baseline / preconditioned).
         \textbf{Ratio} is baseline $\div$ preconditioned ($>1$ means preconditioning helped).
         ``inf'' indicates numerical overflow of the baseline. Best values in each dimension are shown in bold.}
\label{tab:benchmark}
\centering
\setlength{\tabcolsep}{4pt}
\begin{tabular}{lc  rr r  rr r}
\toprule
\multirow{2}{*}{\textbf{Function}}
  & \multirow{2}{*}{$n$}
  & \multicolumn{3}{c}{$\theta$-RMSE}
  & \multicolumn{3}{c}{$f^*$-RMSE} \\
\cmidrule(lr){3-5}\cmidrule(lr){6-8}
& & Base & Pre & Ratio & Base & Pre & Ratio \\
\midrule
\multirow{3}{*}{Scaled Quadratic}
  & 16  & 1.50e-2 & 3.36e-3 & 4.5   & 1.48e-4 & 8.58e-6 & 17.2 \\
  & 64  & 5.57e-1 & 2.09e-2 & 26.6  & 2.32e-1 & 3.55e-4 & 654.7 \\
  & 128 & 2.64e+1 & 2.36e-2 & 1.1e+3 & 7.08e+2 & 2.99e-4 & 2.4e+6 \\
\midrule
\multirow{3}{*}{Quartic Mix}
  & 16  & 8.27e-1 & 3.99e-1 & 2.1 & 5.19e-1 & 1.37e-1 & 3.8 \\
  & 64  & 1.43e+2 & 2.52e+1 & 5.7 & 1.76e+9 & 9.09e+5 & 1.9e+3 \\
  & 128 & inf & inf & -- & inf & inf & -- \\
\midrule
\multirow{3}{*}{Sum Cosh}
  & 16  & 5.57e-1 & 2.73e-1 & 2.0 & 3.82e-1 & 3.86e-2 & 9.9 \\
  & 64  & 8.69e-1 & 5.54e-1 & 1.6 & 4.41e+2 & 5.20e+1 & 8.5 \\
  & 128 & 1.32e+0 & 5.55e-1 & 2.4 & 3.18e+2 & 2.66e+1 & 11.9 \\
\midrule
\multirow{3}{*}{Exp-Quad}
  & 16  & 7.25e-3 & 4.83e-3 & 1.5 & 7.44e-5 & 3.26e-5 & 2.3 \\
  & 64  & 6.35e-1 & 1.16e-1 & 5.5 & 3.82e-1 & 1.29e-2 & 29.7 \\
  & 128 & 7.04e-1 & 2.26e-1 & 3.1 & 3.56e-1 & 3.69e-2 & 9.6 \\
\midrule
\multirow{3}{*}{LogSumExp}
  & 16  & 3.49e-1 & 1.47e-1 & 2.4 & 1.45e-1 & 2.61e-2 & 5.6 \\
  & 64  & 1.75e-1 & 2.09e-1 & 0.8 & 2.75e-2 & 3.36e-2 & 0.8 \\
  & 128 & 2.75e-1 & 1.35e-1 & 2.0 & 1.69e-2 & 1.51e-2 & 1.1 \\
\bottomrule
\end{tabular}
\end{table*}

\subsection{Scope and Limitations}

Despite its generality, Hessian preconditioning carries several
limitations that limit the scope of its applicability.

\textbf{Existence of a global minimizer.}
The method requires a finite minimizer $\theta_0=\operatorname*{argmin}f(\theta)$.
It fails for functions with no global minimum (e.g., $f(\theta)=e^\theta$),
and for functions whose minimizer lies at infinity or is hard to find
numerically.

\textbf{Locality.}
The preconditioning is a second-order approximation valid near $\theta_0$.
For functions whose curvature varies strongly across the domain, the benefit
may not propagate far from the minimizer. This can be seen on the conjugation of the regularized Log-Sum-Exp function (table \ref{tab:benchmark}), where preconditioning does not necessarily reduces RMSE.

\textbf{Hessian conditioning.}
If $H$ is nearly singular (very small eigenvalue), $\Hisq$ amplifies noise and
the preconditioning can be numerically unstable.
The implementation clamps eigenvalues below $\varepsilon=10^{-6}$, but this
degrades the quality of the deformation.

\textbf{When preconditioning is unnecessary.}
If $H\approx I$ (e.g., the function is already paraboloidal), the deformation
has no effect and the overhead is pure cost.
Similarly, for uniformly well-conditioned functions, both methods converge at
comparable speed.

\textbf{Hessian computation cost.}
The row-by-row autograd Hessian computation costs $O(n^2)$ backward passes.
For large $n$, this may be prohibitive; one could resort to a diagonal or
low-rank Hessian approximation, at the cost of a less accurate
preconditioning.

\textbf{Other preconditioning strategies.}
The polarity framework supports broader choices.
Any invertible affine deformation $S$ that makes $f\circ S^{-1}$ easier to
conjugate is valid.
For instance, if prior knowledge about the function's structure is available
(separability, specific spectral profile), a tailored deformation may
outperform the local Hessian approximation.

\section{Conclusion}
\label{sec:conclusion}

We proposed Hessian preconditioning for neural Legendre-Fenchel conjugation,
motivated by the affine invariance of the projective polarity framework.
By deforming the input function so that its local geometry near the minimizer
matches the canonical paraboloid, the conjugation problem in transformed
parameterization becomes close to the identity map, which is the initialization
of the residual network.
The method adds negligible inference overhead (two matrix-vector products per
query) and achieves consistent improvements in convergence speed and
conjugation quality, particularly for ill-conditioned functions.

Future work could explore adaptive preconditioning (updating $\theta_0$
online), learning the deformation jointly with the conjugation network, or
extending the approach to set-valued and stochastic settings motivated by
entropic optimal transport (OT).

\bibliographystyle{plain}
\bibliography{references}

\section{Additional Experimental Results}
\label{app:experiments}

\end{document}